# The Soft Compiler: A Web-Based Tool for the Design of Modular Pneumatic Circuits for Soft Robots

Lauryn Whiteside,[†] Savita V. Kendre,[†] Tian Y. Fan, Jovanna A. Tracz, Gus T. Teran,
Thomas C. Underwood, Mohammed E. Sayed, Haihui J. Jiang, Adam A. Stokes,
Daniel J. Preston, George M. Whitesides, and Markus P. Nemitz[*]

*Abstract*— Developing soft circuits from individual soft logic gates poses a unique challenge: with increasing numbers of logic gates, the design and implementation of circuits leads to inefficiencies due to mathematically unoptimized circuits and wiring mistakes during assembly. It is therefore practically important to introduce design tools that support the development of soft circuits. We developed a web-based graphical user interface, *the Soft Compiler*, that accepts a user-defined robot behavior as a truth table to generate a mathematically optimized circuit diagram that guides the assembly of a soft fluidic circuit. We describe the design and experimental verification of three soft circuits of increasing complexity, using the Soft Compiler as a design tool and a novel pneumatic glove as an input interface. In one example, we reduce the size of a soft circuit from the original 11 logic gates to 4 logic gates while maintaining circuit functionality. The Soft Compiler is a web-based design tool for fluidic, soft circuits and published under open-source MIT License.

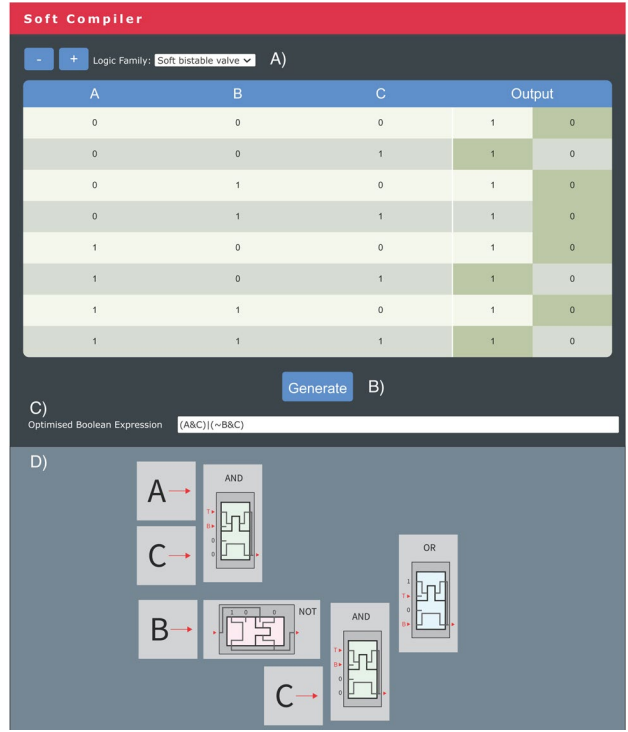

Figure 1. The web-interface of the Soft Compiler. A) The user defines (1) the number of inputs to the soft circuit, (2) the soft logic family (here: "Soft bistable valve") and (3) a truth table, mapping input combinations to a desired output response. B) Once the truth table is defined, "Generate" executes the compiler and outputs an optimized Boolean expression, and D) a schematic depicting the optimal wiring configuration that the user then implements during circuit assembly.

## I. INTRODUCTION

Soft robots, a new class of robots made from elastomeric polymers, have gained significant interest from robotics researchers in recent years [1], [2]. Soft robots possess a wide range of advantages compared to rigid robots including safety and compatibility in interaction with humans and animals, low cost [3], resistance to impact that would damage equivalent hard structures [4], [5], resistance to corrosive chemicals and harsh conditions [6], high cycle lifetime [7], and the ability to incorporate intelligence into its physical structure [3], [4]. Soft robots have become desirable options in a variety of applications including in the assistance of human motor function [8], nursing and elderly care [9], minimally invasive surgery [10], and exploration of terrestrial environments [11], and are typically actuated using pneumatics [12], electrostatics [13], or magnetism [14].

Soft robots currently utilize soft materials for actuation [15], [16], sensing [17], [18], and structure, and rigid components for power, memory, and control [19], [20]. Therefore, the majority of soft robots are tethered to hard components external to the robot (e.g., batteries, pumps, valves, motors, and microprocessors). Efforts to develop fully integrated soft robots have resulted in systems that are hybrids (combinations) of soft and hard components. As the field matures, soft substitutes for electric components have become available. Octobot is a soft robot that performs a pre-programmed function without the use of electronics: it cycles between two different actuation states, resulting in the movement of octopus-like arms [21]. The robot is controlled by a microfluidic, oscillatory circuit and is powered by the catalytic decomposition of an on-board monopropellant fuel supply [21]. Mahon et al. demonstrated a soft state machine composed of fluidic logic gates used to power and switch between vacuum driven actuators [22]. Barlett et al. developed a soft microfluidic demultiplexer that enables the control of multiple pneumatic outputs from only a few (on the order of log n) pneumatic inputs [23].

Of particular importance is work from Rothemund et al., who developed a soft, bistable valve that can be used for the

This work was supported by the Department of Energy, Office of Basic Energy Science, Division of Materials Science and Engineering, grant ER45852, which funded work related to experimental apparatus and demonstrations. We would also like to acknowledge the Harvard NSF funded MRSEC (DMR-1425070) and funding from WPI for salary support of Markus P. Nemitz. Lauryn Whiteside, Savita V. Kendre, Tian Y. Fan, Gus T. Teran, and Markus P. Nemitz are with the Department of Robotics Engineering at Worcester Polytechnic Institute, USA.
Jovanna A. Tracz is with the Eastern Virginia Medical School, USA.

Thomas C. Underwood is with the Department of Aerospace Engineering and Engineering Mechanics at the University of Texas at Austin, USA.
Mohammed E. Sayed and Adam A. Stokes are with the School of Engineering at the University of Edinburgh, UK.
Haihui J. Jiang and George M. Whitesides are with the Department of Chemistry and Chemical Biology at Harvard University, USA.
Daniel J. Preston is with the Department of Mechanical Engineering at Rice University, USA.
† Indicates equal contribution
*To whom correspondence may be addressed: mnemitz@wpi.edu

switching of pressures, representing a soft substitute for an electronic transistor [24]. This non-electronic switch is controlled via pneumatic signals and was used for autonomously gripping an object and controlling the locomotion of an earthworm-like soft robot. Preston et al. reconfigured soft bistable valves and introduced them as fundamental logic units (AND-, OR-, and NOT-gates) that can be assembled to form soft circuits (e.g., digital-to-analog converters and shift registers) and used for the control of soft robots (e.g., an underwater soft robot that repeatedly dives from the surface of water to a predefined depth) [25]. In a subsequent report, Preston et al. also combined three soft, bistable valves, configured as NOT-gates, to create oscillatory circuits which were used in a mechanotherapy device for the lower leg (using sequential inflation and deflation of soft cushions), among other applications [26]. Nemitz et al. further utilized the material properties of soft bistable valves by altering the valve membrane thickness to develop non-volatile memory devices for the permanent storage of information in soft materials [27]. This demonstrated the secure storage of information even after a discontinuation of supply pressure.

Soft bistable valves shown in prior work each have a total of six inputs/outputs (I/Os) [5]; they are modular, and they can be configured as AND-, OR-, or NOT-gates by changing their input-output configurations [25], [28]. We can stack multiple soft valves to implement arbitrary soft circuits within soft robots, that can be used for their control. Soft bistable valves simplify the design of soft circuits: instead of developing monolithic soft circuits, soft bistable valves can be assembled into circuits and subsequently disassembled to be repurposed for new applications. This also permits the replacement of a soft valve that is damaged without requiring reimplementation of the entire circuit.

However, in demonstrations of soft systems controlled using soft bistable valves, circuit design and implementation that was performed manually proved to yield unwieldy results even for relatively simple circuits. Because all logic gates are based on the same soft valve design, they appear the same, and are only discriminable from each other based on their input-output configuration, which presents a practical challenge when implementing complex logic circuitry. This challenge can also be applied to future implementations of multiple, modular soft logic gates, as any modular logic gate will have multiple connections with others to form circuits. Further, the wiring of tubing between logic gates increases in difficulty as the number of valves within the circuit increases. As a soft circuit with 10 soft valves requires up to 60 tubes to be wired, it is easy for human users to face difficulties when configuring circuits.

In this paper, we introduce a web-based Graphical User Interface (GUI), called the Soft Compiler (**Figure 1**), which was released under open-source MIT License and guides the design of pneumatic soft circuits to control soft robotic actuators. The main contributions of this work include:

1. The development of a web-based GUI that allows users to designate a desired soft robotic behavior using a truth table, which the compiler subsequently converts into a schematic diagram.
2. The development of a novel pneumatic glove that serves as an input device to the soft circuits.
3. The demonstration of the Soft Compiler using the pneumatic glove to implement several soft circuits composed of soft, bistable valves.

## II. SOFT COMPILER

### A. Graphical User Interface

The GUI of the compiler allows users to define a truth table (**Figure 1**); the truth table maps fluidic inputs of a soft device (e.g., push buttons) to an output (e.g., a pneumatic actuator). Once the desired robot behavior is defined using the truth table, the compiler outputs an optimized Boolean expression (**Figure 1C**) and image files depicting the wiring schematic (**Figure 1D**). The wiring schematic is dependent upon the soft logic family that was selected (**Figure 1A**). Soft logic families are different physical implementations of soft logic gates and currently the library consists of the soft bistable valve. However, as more roboticists conduct research on fluidic control schemes for soft robots, the library will be extended by the soft robotics community and our research group.

We developed the Soft Compiler as a web-based application that can be accessed at www.roboticmaterialsgroup.com/tools. The GUI uses a heuristic algorithm to generate a schematic and uses the Boolean.py library in Python for logic simplification. We employed Amazon Web Service (AWS) to serve requests for the Boolean algebra simplification using a Gateway API and a Lambda Function. The AWS Lambda function implements the Boolean algebra simplification which is passed by the AWS Gateway API to the Soft Compiler website.

A compiler generally refers to a software tool or a program that converts instructions from a high-level language (e.g., C++) to a lower-level form (e.g., machine code) to create an executable program. We refer to our GUI as a Soft Compiler as it converts a high-level input (in this case a truth table) into a schematic of a low-level implementation of the soft circuit (**Figure 1, Figure 2**). The Soft Compiler is motivated by existing software for electro- and electro-pneumatic circuits. Many software packages have been developed (e.g., FluidSIM, FluidDRAW, SMC PneuDraw) for the design of circuit diagrams that contain mechanical valves, actuators such as cylinders, sensors, and switches. The novelty of the Soft Compiler in comparison to existing software is that it is published under open-source MIT license (i.e., initiated as a community project), web-based (i.e., it can be used on computers and mobile devices alike), and developed for the design and implementation of fluidic, soft circuitry.

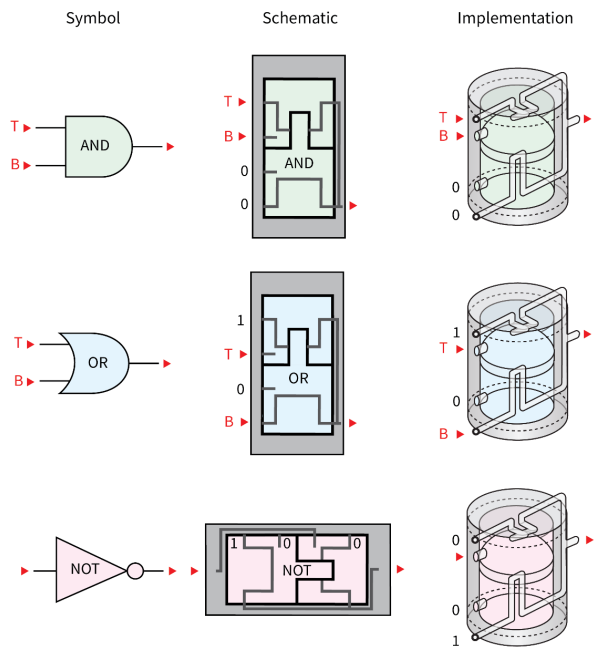

**Figure 2. Abstraction levels**. Generic symbols of logic gates; wiring schematics of logic gates used in the Soft Compiler; representations of implementations of soft bistable valves.

*B. Logic extraction and optimization*

A "truth table" is typically used to identify the relationship between inputs and outputs of a logic circuit [29]. A single representative Boolean expression can be extracted from a truth table using the Sum-Of-Products (SOP) expression. For each row in which the output is binary "1", a Boolean term is formalized. The SOP expression may then be simplified using algorithms for minimization. Optimized circuits require fewer logic gates than non-optimized circuits, which provides a practical advantage for the design and fabrication of soft circuits. Designs that are mathematically unoptimized will suffer from expanding numbers of logic gates requiring fabrication of more devices and evoking additional propagation delays (i.e., time for a circuit to respond to input stimuli). There are several techniques for simplifying Boolean algebra (e.g., using Karnaugh diagrams) [30]. The Soft Compiler uses the Boolean.py library in Python for the extraction and optimization of Boolean expressions from truth tables.

*C. Partitioning and assembly*

After the Boolean logic expression is simplified, it is processed using our partitioning algorithm to generate the wiring schematic for a soft circuit. Our partitioning algorithm, implemented in JavaScript and displayed using HTML and CSS, proceeds in two key steps: (1) expression blocks are broken into sub-blocks, and (2) logic elements are positioned in different windows using a gridded coordinate system. Each sub-block is saved as a different image and referenced in the base level using the sub-block number. The partitioning algorithm divides the expression by OR-gates to create sub-expressions. These sub-expressions are divided by AND-gates to yield input letters or their NOT counterparts. Then, the letters are mapped to their corresponding image such that the logic schematic can be visualized. These letters are always located to the far left of the schematic since they are the input to the expression. If the input letter is negated, then a corresponding image to the NOT-gate is appended to its right. The NOT-gate image always accepts a single input and returns a single output. Subsequently, the corresponding image of logic operators (i.e., AND, OR, etc.) that hold together the sub-expressions accept two inputs and returns a single output. These logic operators are appended to the right of its sub-expressions and aligned such that its two inputs are evident. This process is repeated until the complete wiring schematic

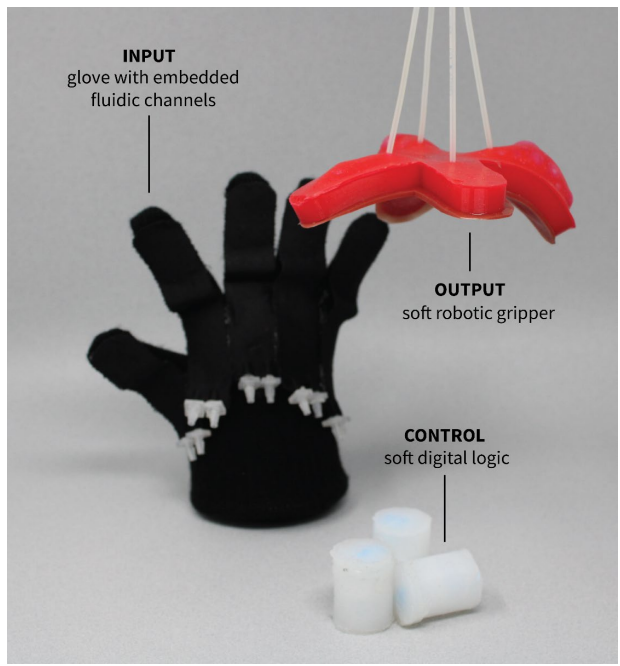

**Figure 3. Soft robotic components used to test Soft Compiler.** Glove with integrated fluidic channels acts as a sensor interface; soft gripper with four individually actuated legs acts as an output; and soft bistable valves configured as soft digital logic gates act as control elements.

is generated. The algorithm can be generalized to any number of inputs.

Graphical representations will vary depending on the soft logic family that is selected for use in the soft circuit. Soft logic family refers to the physical logic gates that constitute the circuit. In this example, the soft logic family depicts the soft bistable valve **(Figure 2)**, which is selected in the GUI **(Figure 1A).** For future implementations of logic gates, we recommend authors design graphical representations of their logic devices too, so the library within the compiler can be extended, and their devices easily used.

In the case of soft bistable valves, schematic representations of the logic gates possess red arrows, where some are marked with "T" and "B", denoting top and bottom respectively **(Figure 2)**. These arrows indicate the connection points between adjacent logic gates. AND- and OR-gates have two inputs and one output, and NOT-gates have one input and one output. The general rule that applies during the connection of arrows within the wiring schematic is as follows: the output arrow of a logic gate always connects to the input arrow of a subsequent logic gate according to its spatial position within the schematic **(Figure 1D)**.

The soft bistable valve can be configured as either an OR-gate, an AND-gate, or a NOT-gate. These fundamental logic gates depict the most efficient implementations of the soft bistable valve (one soft valve per basic logic gate). Any other logic function must be assembled from multiple soft bistable valves. New fluidic logic devices that operate at different physical principles will require modifications to the compiler. For example, if a single logic device had the functionality of a NAND-gate, the most efficient implementations of soft circuits may be *specific* combinations of NAND-gates, and therefore the compiler would require a different output response than for the soft bistable valve logic family. Therefore, we included soft logic families, to account for different physical principles of future logic gates.

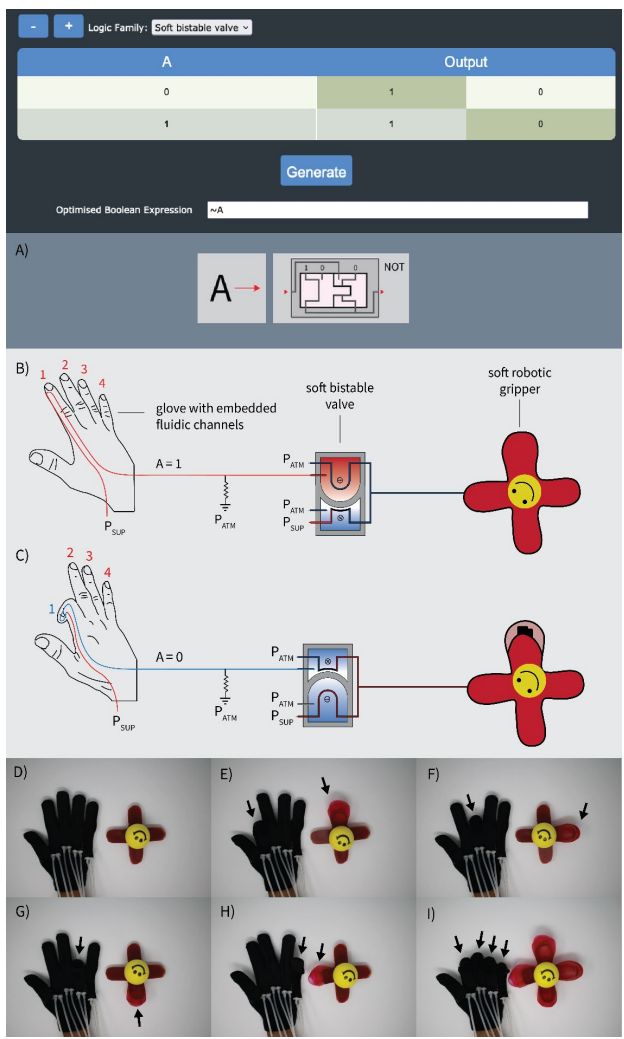

**Figure 4. Direct mapping.** A) When a finger is bent, the corresponding glove output is low. The truth table maps this binary "0" input, to a binary "1" output, so that each finger can control a different leg of the soft robotic gripper. The schematic indicates that the output of our glove has to be connected to a soft bistable valve that is configured as a NOT-gate, and then connected to the soft gripper leg. B) Supply pressure is applied to each finger of the glove. C) When a finger is bent, the supply pressure to that finger is interrupted, and a pull-down resistor depletes the trapped pressure. D-I) Each finger is connected to a soft bistable valve configured as a NOT-gate, and when a finger is bent, a leg of the soft gripper is being inflated.

## III. PNEUMATIC GLOVE

### A. Operating principle

The development of soft circuits, their demonstration, and comparison with other logic implementations requires generic input and output devices. While there is a great variety of pneumatic actuators, fluidic inputs or sensory devices are less common. To that end, we developed a pneumatic glove with embedded fluidic channels (**Figures 3 and 4**); this fluidic input device can theoretically generate 16 ($2^4$) different input combinations (pinky, ring finger, middle finger, and index finger). When a finger is bent, the supply pressure is cut off and the trapped air is depleted through a pull-down resistor to atmospheric pressure (**Figure 4B/4C**), thus the output of the finger is LOW (**A = 0**). When the finger is not bent, the supply pressure can pass through the glove and the output of the finger is HIGH (**A = 1**).

### B. Materials and fabrication process

The tubing of the pneumatic glove was made from two thermoplastic polyurethane (TPU) coated textile patches. We cut and placed parchment paper in U-shape between the textile patches and used a heat press at 220 °C for 35 seconds to bond the TPU coated sides of the textile. We used a heat press from Fancierstudio and a Cameo 4 blade plotter. We fabricated and attached four such textile tubes to a wool glove using hot glue. Each textile tube has two connectors attached to it; we connected supply pressure to one connector, and a pull-down resistor of 30 cm length and an inner diameter of 0.5 mm to the other connector.

**Table 1. Bill of Materials**

| Description | Supplier | Unit | Cost |
|---|---|---|---|
| Textile (coated Nylon Taffeta) | Seattle Fabrics Inc. | 1 inch | $0.02 |
| Parchment paper | Amazon.com | 1 ft | $0.06 |
| Wool glove | Amazon.com | 1 | $1.50 |
| **Total** | | | **$1.58** |

## IV. DEMONSTRATION

To demonstrate the circuit design process using the Soft Compiler, we present three distinct soft circuits that map the pneumatic glove with a soft robotic gripper. The glove serves as an input interface, the gripper as an output interface, and soft bistable valves as logic gates (**Figure 3**). We provide a video in the Supplemental Information that demonstrates the use of the GUI of the Soft Compiler and the experimental validation of the soft circuits (**Video S1**).

### A. Direct mapping

Direct mapping, or one-to-one mapping, demonstrates the ability of a user to control each leg of a soft gripper with a different finger. This demonstration is sometimes called the *digital twin*, as the soft robot mimics the behavior of the human user. To achieve this functionality, we created a truth table that reflects our input and output devices. Since each finger that is bent in the glove outputs LOW (**A = 0**), and each leg of our robotic gripper inflates when receiving HIGH (**Q = 1**), we inserted this inverse behavior into the truth table (**Figure 4A**). The Soft Compiler generates a schematic that instructs the user to: (i) connect the output of a finger from the pneumatic glove to the input of a soft bistable valve that is configured as a NOT gate; (ii) and to connect the output of the NOT gate to an input of the soft robotic gripper (**Figure 4B-C**). Since each finger has the same functionality but is connected to a different leg of the soft robotic gripper, we implement four NOT gates to connect the entire glove to the soft gripper. Any combination of fingers bent results in the actuation of corresponding legs of the soft robotic gripper (**Figure 4D-I**).

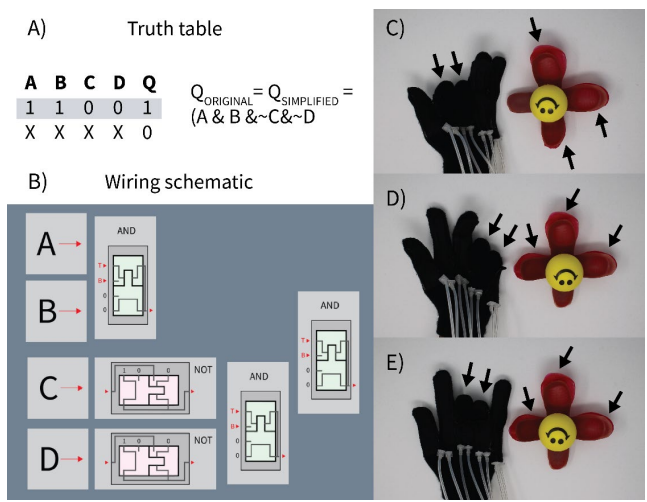

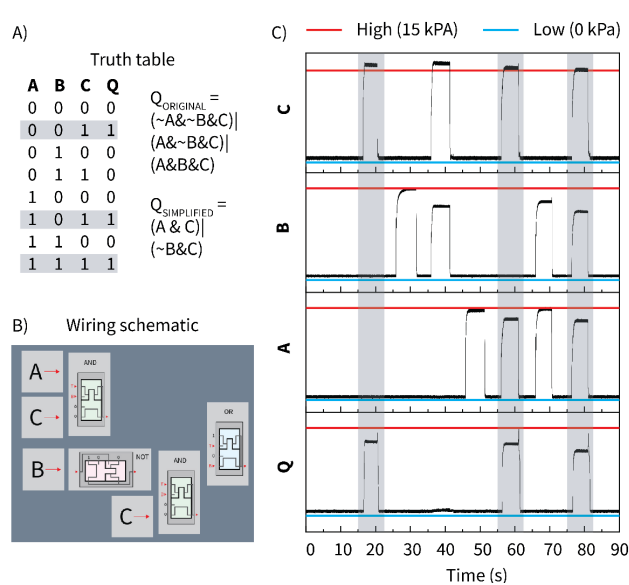

**Figure 5. Two finger mapping.** A) Truth table of the input combination that results in an output response. B) Wiring schematic that indicates how to connect the glove and the gripper with the soft circuit. C-E) Three different implementations of the same circuit that lead to the actuation of the soft gripper.

*B. Two-finger mapping*

We also demonstrate a soft circuit that only responds to a specific combination of inputs, rather than to any combination as shown in the previous demonstration. When inputs A and B are HIGH (**A** = **B** = **1**), and inputs C and D are LOW (**C** = **D** = **0**), only then the soft circuit outputs HIGH (**Q** = **1**) (**Figure 5A**). The output of the soft circuit can be attached to an arbitrary number of pneumatic outputs, and in our case, we connected the output to three legs of the soft robotic gripper (**Figure 5C-E**). We also demonstrate different combinations of two-finger inputs resulting in an output. The four inputs to the soft circuit can be arbitrarily changed, so that inputs C and D, previously being index and middle fingers (**Figure 5C**), are now ring finger and pinky (**Figure 5D**), or middle and and ring fingers (**Figure 5E**). Note that our glove outputs LOW when a finger is bent, and therefore inputs indicating "0" in the truth table drive the output response of the soft circuit.

*C. Complex mapping*

The final demonstration showcases the importance of implementing mathematically optimized soft circuits (**Figure 6**). We created a truth table with three inputs. When inputs A and B are LOW (**A** = **B** = **0**), and input C is HIGH (**C** = **1**), or when inputs A and C are HIGH (**A** = **C** = **1**), and input B is LOW (**B** = **0**), or when all three inputs are HIGH (**A** = **B** = **C** = **1**), the output is HIGH (**Q** = **1**). The SOP expression extracted from the truth table is:

$$Q = (\sim A \& \sim B \& C) | (A \& \sim B \& C) | (A \& B \& C)$$

If we implemented this soft circuit, we would require 6 AND-gates, 3 NOT-gates, and 2 OR-gates, for a total of 11 soft bistable valves. The soft bistable valve is a mechanical device, that flips a membrane to switch airflow, and therefore, the more logic gates are implemented in a circuit, the longer the output response takes. This is particularly true for logic gates that are connected in series, as each valve waits for its adjacent valve to reach saturation. We refer to work on soft digital logic for an in-depth analysis of propagation delays of soft bistable valves [5]. The optimized Boolean expression for the Soft Compiler is:

$$Q = (A \& C) | (\sim B \& C)$$

This soft circuit only requires 2 AND-gates, 1 NOT-gate, and 1 OR-gate, for a total of four soft bistable valves. The implementation of the optimized Boolean expression requires fewer valves than the unoptimized expression. We assembled and tested the soft circuit for the optimized implementation. The truth table and the output response of the soft circuit match one another (**Figure 6**).

**Figure 6. Complex mapping.** A) The truth table shows three arbitrary combinations of inputs that result in an output response. The optimized term requires fewer logic gates than the unoptimized term. B) Schematic that instructs users how to connect inputs with logic gates, and logic gates with outputs. C) Experimental verification of the soft circuit. Truth table and measurements match one another.

## V. DISCUSSION

The library of soft logic families currently only includes soft bistable valves. This limits the useability of the Soft Compiler to researchers with access to a soft robotics laboratory to fabricate soft bistable valves. As the community continues to develop new types of soft logic families, they will be added to the library of the Soft Compiler.

The Soft Compiler only outputs wiring schematics at this stage. In future versions, we will add a performance report to generated soft circuits that shows critical parameters based on the physics of the soft logic family such as the maximum propagation delay and how many logic gates were removed through optimization.

The soft compiler only generates combinational (or memoryless) logic circuits. Combinational logic allows for mostly reactive robot behaviors in which input stimuli generate output responses. In future work, we plan on developing a GUI that abstracts from the gate level and allows for high-level implementations including volatile and non-volatile memory elements. To that end, we also envision a process similar to Design for Manufacturing (DFM) that assesses a design of a soft circuit for potential issues such as maximum supply pressure required, propagation delay, and cumulative size of soft circuit.

Because the Soft Compiler can be readily used as a web-application, we plan on integrating the compiler in a classroom activity and with the goal to investigate the benefits and limitations of the compiler and determine future iterations. Overall, this work depicts the starting point of a larger effort towards the development of design tools for roboticists creating fluidic circuits to control soft robots.

## VI. CONCLUSION

We developed a web-based GUI, *the Soft Compiler*, that accepts a user-defined robot behavior as a truth table to generate a mathematically optimized circuit diagram that guides the assembly of a soft circuit. In our example, circuits are based on soft bistable valves configured as NOT-, AND-, and OR-gates. We demonstrated three different soft circuits with increasing complexity and a novel pneumatic glove as an input interface. The development of software tools that support the creation of soft robotic systems including soft actuators, sensors, and controllers will become increasingly important. Not only will they facilitate the creation of soft robotic systems, but also simplify their implementation.